\pdfoutput=1

\documentclass[11pt]{article}

\usepackage[final]{ACL2023}

\usepackage{times}
\usepackage{latexsym}
\usepackage{amsmath}
\usepackage{amsfonts}
\usepackage{amssymb}
\usepackage{pifont}
\usepackage[ruled,linesnumbered]{algorithm2e}
\DontPrintSemicolon

\usepackage{amsmath}  
\usepackage{natbib}
\usepackage{booktabs}

\usepackage{listings}
\usepackage[T1]{fontenc}

\usepackage[utf8]{inputenc}

\usepackage{microtype}
\usepackage{graphicx}

\usepackage{inconsolata}
\usepackage{multirow}
\usepackage{makecell}

\title{Encouraging Good Processes Without the Need for Good Answers: Reinforcement Learning for LLM Agent Planning}

\author{
    Zhiwei Li\textsuperscript{1}\thanks{ \quad Work was done when the authors were interning at WeChat AI, Tencent Inc., China.},
    Yong Hu\textsuperscript{1}, 
    Wenqing Wang\textsuperscript{1,2}\footnotemark[\value{footnote}] \\
    \textsuperscript{1} WeChat, Tencent Inc., China \\
  \textsuperscript{2} School of Software \& Microelectronics, Peking University, Beijing \\
    \texttt{zhiweili.jay@foxmail.com, rightyonghu@tencent.com, wangwenqing@stu.pku.edu.cn}
}

\begin{document}
\maketitle
\begin{abstract}
The functionality of Large Language Model (LLM) agents is primarily determined by two capabilities: action planning and answer summarization. The former, action planning, is the core capability that dictates an agent's performance. However, prevailing training paradigms employ end-to-end, multi-objective optimization that jointly trains both capabilities. This paradigm faces two critical challenges: imbalanced optimization objective allocation and scarcity of verifiable data, making it difficult to enhance the agent's planning capability. To address these challenges, we propose Reinforcement Learning with Tool-use Rewards (RLTR), a novel framework that decouples the training process to enable a focused, single-objective optimization of the planning module. Crucially, RLTR introduces a reward signal based on tool-use completeness to directly evaluate the quality of tool invocation sequences. This method offers a more direct and reliable training signal than assessing the final response content, thereby obviating the need for verifiable data.  Our experiments demonstrate that RLTR achieves an 8\%–12\% improvement in planning performance compared to end-to-end baselines. Moreover, this enhanced planning capability, in turn, translates to a 5\%–6\% increase in the final response quality of the overall agent system.
\end{abstract}

\section{Introduction}
Large Language Models (LLMs) have achieved significant advancements in natural language processing, including code generation~\cite{wang2023review}, question answering~\cite{shailendra2024survey}, and reasoning~\cite{wei2022chain}. These breakthroughs have spurred interest in developing agents based on LLMs~\cite{cheng2024exploring,shen2024llm}. A typical agent workflow consists of two main stages: the planning stage, in which tool calls are made to gather information, and the summary stage, where the collected information is synthesized to generate the final response~\cite{wang2024survey}. Between the two, the planning stage is crucial in the agent system. The accuracy of the agent's final output heavily depends on the comprehensive information collected through complete tool calls.

\begin{figure}[t]
    \centering
    \includegraphics[width=0.8\linewidth]{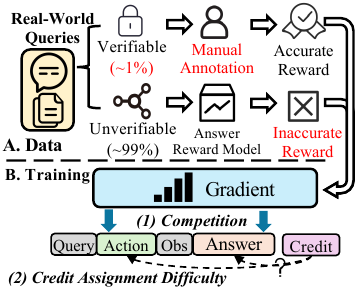}
    \caption{Two main challenges in end-to-end agent training for industry scenarios: (A) Lack of effective rewards for predominant data; (B) Optimization competition and difficulties in credit assignment.}

    \label{fig:example}
\end{figure}

Currently, the dominant end-to-end reinforcement learning (RL) paradigm for agents in LLMs performs multi-objective optimization for both the planning and summary policies. It uses the final summary's answer as the reward to update both policies simultaneously~\cite{yang2025qwen3, glm2024chatglm}. Integrated optimization effectively enables end-to-end LLMs to provide comprehensive agent capabilities. However, tightly coupled multi-objective RL optimization presents data and training challenges when fine-tuning for industrial business scenarios, as outlined below:

\noindent\textbf{Challenge 1: Lack of Effective Rewards for Predominant Data.} As illustrated in Figure~\ref{fig:example}(A), current advanced end-to-end agent RL methods~\cite{jin2025search,feng2025retool,li2025torlscalingtoolintegratedrl} rely on training data with verifiable answers to compute accurate rewards. However, in real-world scenarios, such data is scarce and requires costly manual annotation~\cite{wu2025webdancer}. For the vast majority of non-verifiable data, a reward model based on the final answer is typically used to validate the output~\cite{yang2025qwen3,guo2025deepseek}, which is prone to reward hacking~\cite{gao2024designing}, leading to inaccurate rewards. As a result, most data lacks effective RL optimization methods.

\noindent\textbf{Challenge 2: Competing Objectives and Credit Assignment Difficulty.} As depicted in Figure~\ref{fig:example}(B), the gradients for the planning and summary modules are often in opposition, making it non-trivial to balance their respective objectives. This issue is further exacerbated in RL, where the reward structure is tightly coupled: the evaluation of the final summary determines the reward for the entire trajectory, which in turn guides the end-to-end policy update. Such a mechanism results in difficult credit assignment~\cite{nguyen2018credit}, whereby correct actions within the trajectory may be unduly penalized for errors originating in the final response. Ultimately, this impedes the optimization of planning capabilities.

The challenges arise from multi-objective optimization that aims to improve both planning and summarization. To address this, we focus solely on optimizing the agent's core planning component (Planner), simplifying the task into a single-objective optimization and mitigating issues related to competing objectives and credit assignment (Challenge 2). To tackle this focused optimization problem, we propose the Reinforcement Learning with Tool-use Rewards (RLTR) framework. We initialize the Planner using knowledge distillation and rejection sampling, then replace the complex final-answer correctness reward with the simpler, more reliable tool-use completeness reward. This reward focuses solely on the action sequence, eliminating the need for final answer verification and addressing data scarcity issues (Challenge 1). The Planner is subsequently optimized using these completeness rewards within a multi-turn RL environment. The optimized Planner is modular and can be paired with any LLM as a summarizer to form a complete agent. Experimental results show that an Planner trained via RLTR improves action performance by 8\%–12\% compared to its end-to-end trained counterpart. Without training a dedicated summary component, this enhancement in the planning stage still leads to a 5\%–6\% improvement in the agent’s end-to-end response performance.

Our contributions can be summarized as follows:
\begin{itemize}
    \item We analyze the challenges inherent in applying end-to-end optimization for agents in industrial scenarios, and propose a targeted single-objective paradigm that focuses on optimizing the agent’s core planning component.
    \item We design a novel reward function based on tool-use completeness, which provides a high-fidelity score for action quality and effectively addresses the challenge of insufficient reward signals in reinforcement learning for the majority of data in industrial scenarios.
    \item Our approach enables stable and effective training of the Planner during both the supervised fine-tuning (SFT) and reinforcement learning (RL) phases, yielding an 8\%–12\% improvement in planning performance. We further demonstrate that enhancing the agent’s planning capability benefits the overall system, translating into a 5\%–6\% average increase in end-to-end response accuracy.
\end{itemize}

\section{Problem Formulation}
We define the multi-objective optimization for an end-to-end agent and the single-objective optimization focusing on actions. In both cases, the task is modeled as a sequential decision process, where a single interaction for a given query is represented as a trajectory $\tau = (s_0, a_0, \dots, s_T, a_T)$, with $T$ being the termination step. The state includes the query $q$ and the history of tool interactions $H_t$, and the action space consists of $K$ tools from $\mathcal{T}$ and the terminal action ANSWER.

The optimization objective for the end-to-end agent $\pi_{e}$ employs the final answer reward function $R_{e}$ and integrates planning with summary generation, and is defined as follows:

\[
\pi_{e}^* = \arg\max_{\pi_{e}} \mathbb{E}_{\tau \sim \pi_e} R_{e}(\pi_e(a_T),y^*)
\]

Our approach optimizes the Planner policy $\pi_p$ using an action planning score function $R$. Once the Planner is sufficiently optimized and outputs the planning trajectory, the Summarizer $\pi_s$ generates the final end-to-end response $y$. The overall process is formally defined as follows:

\begin{equation}
\begin{aligned}
 \pi_p^* &= \arg\max_{\pi_p} \mathbb{E}_{\tau \sim \pi_p} [R(\tau)],\\
y &= \pi_s(\tau), \text{where} \quad \tau \sim \pi_p^*.\\
\label{eq:optimal_policy}
\end{aligned}
\end{equation}

\begin{figure*}[t]
    \centering
    \includegraphics[width=1\linewidth]{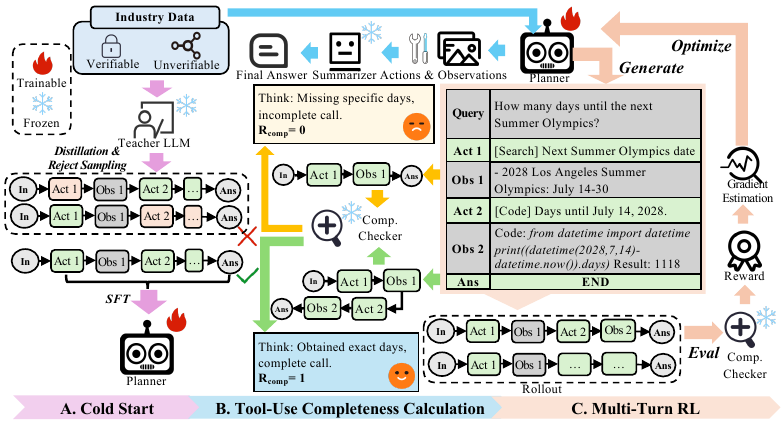}
    \caption{Our RL with Tool-use Rewards (RLTR) framework. Initially, we perform (A) knowledge distillation and rejection sampling to cold-start the Planner. In (B), we compute the tool-use completeness reward for the Planner's action sequence using an existing LLM. Finally, in (C), we optimize the Planner's tool use through multi-turn reinforcement learning. The complete training template corresponding to the example in (C) can be found in Appendix~\ref{app:template}. ``Comp.'' denotes ``Completeness.''}
    \label{fig:framework}
\end{figure*}

\section{Framework}
We begin by performing cold-start initialization of the Planner using knowledge distillation and rejection sampling. Subsequently, we introduce the calculation of tool-use completeness, which serves as the primary reward signal, and employ multi-turn reinforcement learning to optimize the Planner. Finally, we utilize a LLM as a Summarizer, which receives the Planner's plan and the corresponding information to generate the final end-to-end response.

\subsection{Cold Start} 
\label{sec:dsft} 
As shown in Figure~\ref{fig:framework}(A), we utilize a state-of-the-art LLM as the teacher model to perform knowledge distillation~\cite{gou2021knowledge} for cold-starting the Planner. Initially, we sample multiple agent action trajectories from the teacher LLM by providing agent simulation commands and questions as input. Subsequently, we apply rejection sampling using the same teacher model to select the trajectories, retaining the best-of-n as the training data. We use the question as input and the teacher LLM's action trajectories as output to perform supervised fine-tuning (SFT) of the Planner for cold-starting, thereby enhancing the model's ability to handle the latest planning task formats.

\subsection{Tool-Use Completeness Calculation}
For most industrial data with unverifiable outcomes, validation requires assessing both action sufficiency and summary accuracy, complicating reward reliability. \textit{Figuring out if something can be done is easier; ensuring it’s done correctly is harder.} By shifting the reward focus to the Planner, we decouple these factors and directly evaluate action sequence integrity, enabling simpler and more effective agent assessment (see Section~\ref{sec:reward_func}). To formalize this, we introduce a completeness-checking function $\mathcal{\gamma} : \mathcal{S} \to \{0, 1\}$, where $\mathcal{\gamma}(s)$ indicates whether the action sequence at state $s$ is complete (1) or incomplete (0). This function is implemented using a verification LLM (Comp. Checker) with a check instruction, as detailed in Appendix~\ref{app:prompt_check}. The invocation completeness is computed by averaging the results over $N$ samples, as follows:
\begin{equation}
\begin{aligned}
R_{comp} = \frac{1}{N}\sum_{i=1}^{N} \gamma_i(\tau)
\label{eq:completeness}
\end{aligned}
\end{equation}

\subsection{Multi-Turn Reinforcement Learning}
We first compute the overall tool-use reward, denoted as $R_{total}$. Initially, we verify whether the agent’s trajectory format is correct. If the format is incorrect, we immediately assign a reward of $-1$; otherwise, we proceed to calculate the tool-use completeness reward $R_{comp}$. Additionally, we incorporate rule-based rewards to stabilize training, including a negative repetition reward $R_{repeat}$ to discourage redundant tool calls, and a negative reward $R_{error}$ as a penalty for incorrect tool usage. These negative rewards are aggregated as $R_{rule} = R_{repeat} + R_{error}$. The total reward is defined as follows:
\begin{equation}
\begin{aligned}
R_{total} = \begin{cases}
-1, \text{if trajectory format is invalid}, \\
R_{comp} + R_{rule}, \text{otherwise}.
\end{cases}
\end{aligned}
\end{equation}

Then, we employ this tool-use reward function for scoring. During multi-turn template construction, we mask the tool-use results to exclude them from the loss computation. This prevents gradient signal dilution and ensures that the agent remains focused on optimizing tool invocation behavior. The detailed template construction process is provided in Appendix~\ref{app:train_template}. We further refine the Planner's optimization objective (Equation~\ref{eq:optimal_policy}) to the standard RL formulation:
\begin{equation}
\begin{aligned}
\pi_p^* = \arg\max_{\pi_{p}} \mathbb{E}_{x\sim\mathcal{D}, a\sim\pi_{p}(\cdot|x;\mathcal{T})}\left[R(x,a)\right] \\- \beta\mathbb{D}_{\mathrm{KL}}\left[\pi_{\theta}(a|x;\mathcal{T}) | \pi_{\mathrm{ref}}(a|x;\mathcal{T})\right]
\end{aligned}
\end{equation}
where $x$ denotes data sampled from distribution $\mathcal{D}$, a indicates tool-use actions sampled from the policy, $\beta$ controls the KL regularization strength, and $\pi_{\mathrm{ref}}$ is the reference model. We optimize this objective through an online RL process. First, the current policy model $\pi_{a}$ generates a batch of tool-use trajectories, which are then evaluated using the action reward function $R$. First, the reward signal is used to compute value estimates, and these estimates then guide policy updates to increase the likelihood of high-reward behaviors. Consequently, this ``Generate-Evaluate-Optimize'' loop is repeated until policy convergence. The complete algorithm is detailed in Appendix~\ref{app:rl_alg}.

Finally, we construct the complete agent pipeline. Either a trained or untrained LLM can be used as the Summarizer $\pi_s$. The input query is first passed to the optimized Planner $\pi_p^*$, which uses tools to collect information and form a trajectory $\tau \sim \pi_p^*$. This trajectory is subsequently input into the Summarizer, which generates the final response: $y = \pi_s(\tau)$.
\section{Experiments}

\begin{table*}[t]
    \centering
    \begin{tabular}{cccccccccc}
        \hline
        \toprule
        \multicolumn{3}{c}{\multirow{2}{*}{}} & \multicolumn{6}{c}{\textbf{Industry}} & \multicolumn{1}{c}{\multirow{2}{*}{\makecell{\textbf{Open-}\\\textbf{source}}}} \\ \cline{4-9} 
        \multicolumn{3}{c}{} & \multicolumn{3}{c}{\textbf{Normal}} & \multicolumn{3}{c}{\textbf{Hard}} & \multicolumn{1}{c}{} \\
        \hline
        \textbf{Model} & \textbf{PO} & \textbf{Method} & \textbf{Com.} & \textbf{Hel.} & \textbf{Rel.} & \textbf{Com.} & \textbf{Hel.} & \textbf{Rel.} & \textbf{Match} \\
        \hline
        Qwen3-235B& \ding{56} & DIRECT & 67.2 & 71.5 & 72.7 & 50.5 & 47.4 & 46.4 &45.8 \\
        DeepSeek-R1 & \ding{56} & DIRECT & 68.8 & 71.2 & 76.0 & 49.7 & 57.5 & 51.5&49.5 \\
        \hline
        \noalign{\vskip 0.1pt}
        \multirow{5}{*}{Qwen3-1.7B}
        & \ding{56} & DIRECT & 44.9 & 41.9 & 46.7 & 22.4 & 30.4 & 32.5 & 29.8 \\
        & \ding{56} & E2E SFT & 56.3 & 59.2 & 62.4 & 30.1 & 35.3 & 37.7 & 37.1 \\
        & \ding{52} & SFT & 60.1 & 61.3 & 64.5 & 35.3 & 38.6 & 41.4 & 39.4 \\
        & \ding{56} & E2E RL & 62.4 & 63.5 & 65.6 & 37.5 & 41.4 & 45.2&40.0 \\
        & \ding{52} & RLTR(Ours) & \textbf{70.2} & \textbf{68.4} & \textbf{72.6} & \textbf{45.4} & \textbf{48.4} & \textbf{49.4}&\textbf{45.6} \\
        \hline
        \noalign{\vskip 0.1pt}
        \multirow{5}{*}{Qwen3-8B}
        & \ding{56} & DIRECT & 51.5 & 53.8 & 65.2 & 35.3 & 36.3 & 37.4 & 35.3 \\
        & \ding{56} & E2E SFT & 66.0 & 65.4 & 70.2 & 40.4 & 45.5 & 44.8 & 41.4 \\
        & \ding{52} & SFT & 67.2 & 70.1 & 71.3 & 46.4 & 48.4 & 51.4&44.4 \\
        & \ding{56} & E2E RL & 69.6 & 71.2 & 76.7 & 44.4 & 47.4 & 53.5&45.2 \\
        & \ding{52} & RLTR(Ours) & \textbf{82.7} & \textbf{76.7} & \textbf{80.9} & \textbf{54.5} & \textbf{61.6} & \textbf{65.6}&\textbf{51.6}  \\ \hline 
        \toprule
    \end{tabular}
    \caption{Performance of different models using various optimization methods. ``PO'' denotes optimization of the Planner only, not the end-to-end agent (including both planning and summary). ``DIRECT'' indicates directly using the original LLMs for tool calls and answers. The values represent percentages, with the ``\%'' symbol omitted. \textbf{Bolded} values indicate the optimal performance for the 1.7B and 8B models. }
    \label{tab:main}
\end{table*}

Similar to DeepResearch~\footnote{https://openai.com/index/introducing-deep-research/}, our environment integrates both a search tool for information retrieval and a code tool for computational tasks, with detailed tool descriptions provided in Appendix~\ref{app:tool}. 

\subsection{Datasets} 
The datasets we use include both in-house industry datasets for training and testing, and open-source datasets as additional test sets.

\noindent \textbf{Industry Dataset} We have collected an industry-level Chinese agent dataset supporting training on both search and code tools. The dataset includes approximately 4k training samples and 0.5k test samples, with the test set categorized into normal and hard levels, providing a rigorous challenge for assessing action performance on difficult queries.

\noindent \textbf{Open-source Dataset} Since our focus is on Chinese scenarios, we utilize the ChineseSimpleQA~\cite{he2024chinese} open-source Chinese QA dataset for additional performance evaluation. To rigorously assess the agent's capabilities, we exclude samples that can be correctly answered by DeepSeek-R1~\cite{guo2025deepseek} or Qwen3-235B-A22B~\cite{yang2025qwen3} without tool invocation. This filtering results in a challenging test set of 855 samples that require tool invocation for accurate resolution.

\subsection{Baselines}
We selected the leading LLMs in the agent field, Qwen3-235B-A22B and DeepSeek-R1, for testing and comparison. We also included commonly used end-to-end agent optimization approaches. One is supervised fine-tuning of both actions and answers (E2E SFT). The other is reinforcement learning, where the final answer serves as the reward (E2E RL), as in Qwen3~\cite{yang2025qwen3}. To ensure a fair comparison, all models were trained and tested within the same interaction environment (see Appendix~\ref{app:env}) and on the same dataset. 

\subsection{Experiment Setup}
\noindent \textbf{Training} In the SFT stage, the end-to-end agent incorporates the answer into the trajectory, while the decoupled agent independently fine-tunes the Planner using only actions. In the RL stage, we follow RLAIF~\cite{leerlaif} and use the original Qwen3-30B-A3B~\cite{yang2025qwen3} as the scoring model to calculate rewards. E2E RL computes the reward based on the correctness of the final response using answer verification instructions (Appendix~\ref{app:answer_prompt}), while RLTR uses tool-completeness verification instructions (Appendix~\ref{app:prompt_check}) to calculate the reward for the Planner’s tool completeness. Further implementation details are provided in Appendix~\ref{app:detail}.

\noindent \textbf{Evaluation} Planning evaluation focuses solely on the action sequence itself. For final response evaluation, we consider two settings: (1) for the end-to-end agent, the final response state is selected as the answer; (2) for the Planner, the original Qwen3-1.7B and Qwen3-8B models serve as summarizers, receiving all actions and observations generated by the Planner as input, and their output is used as the answer for evaluation. All metrics are computed using the Qwen3-235B-A22B model.

\subsection{Metrics}
\noindent \textbf{Planning Metric} Tool-use completeness (Com.), as defined in Equation~\ref{eq:completeness}, is employed as the evaluation metric for planning.

\noindent \textbf{Final Response Metrics} For open-source data with ground-truth answers, we use the matching accuracy (Match) between the ground-truth answers and the agent-generated responses to evaluate the quality of the final response. For industry data lacking standard answers, we use common metrics such as helpfulness (Hel.) and relevance (Rel.)~\cite{guo2023evaluating} to assess the final response.

\subsection{Main Results}
\begin{figure}[htbp]
    \centering
    \includegraphics[width=1\linewidth]{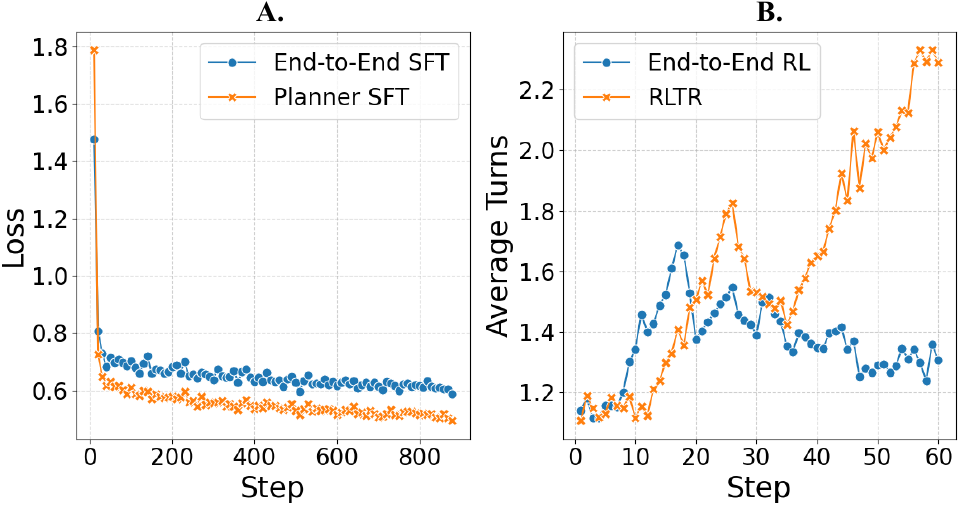}
    \caption{(A) Loss trends and (B) average number of turns for the end-to-end agent and Planner during the SFT and RL training stages.}
    \label{fig:sft_loss}
\end{figure}
\noindent \textbf{Planning Performance Results} As shown in Table~\ref{tab:main}(Industry), the Planner demonstrates superior planning optimization performance compared to the end-to-end agent in both the SFT and RL stages for the 1.7B and 8B models, with the primary tool-use completeness metric exhibiting an average improvement of approximately 8\%–12\%. This improvement is particularly pronounced on the hard subset, underscoring the critical importance of targeted planning capability optimization for addressing complex problems. This improvement can be further explained by the loss and average number of turns observed during the training process. As shown in Figure~\ref{fig:sft_loss}(A), during the SFT phase, the Planner achieves faster convergence and lower loss than the end-to-end agent, reflecting more effective gradient optimization of the planning process. In the RL phase (Figure~\ref{fig:sft_loss}(B)), the Planner with RLTR more effectively activates the LLMs to utilize tools, while end-to-end RL fails to do so. This advantage arises from RLTR’s precise reward attribution to Planner actions, whereas end-to-end RL, constrained by final answer rewards, struggles with effective credit assignment.

\noindent \textbf{Final Response Performance Results} As shown in Table~\ref{tab:main}, the Planner outperforms the end-to-end agent in final response performance in both the SFT and RL stages across various models, with an average improvement of approximately 5\%-6\%. This suggests that the enhanced planning performance of the Planner contributes to the overall improvement of the complete agent. Specifically, Planner optimization enables more effective tool use and comprehensive information gathering, thereby improving Summarizer accuracy even without dedicated summarization training.

\begin{table}[htbp]
    \centering
    \begin{tabular}{cccc}
        \hline
        \toprule
        \textbf{Method} & \multicolumn{2}{c}{\textbf{Industry}} & \multicolumn{1}{c}{\multirow{2}{*}{\makecell{\textbf{Open-}\\\textbf{source}}}} \\ \cline{2-3}
        & \textbf{Normal} & \textbf{Hard} & \\
        \hline 
        E2E PPO & 69.6 & 44.4 & 45.2\\
        \hline 
        GRPO & 76.4 & 49.3 & 47.4\\
        REINFORCE++ & 82.1 & 52.5 & 50.3\\
        PPO & \textbf{82.7} & \textbf{54.5} & \textbf{51.6}\\
        \hline
        \toprule
    \end{tabular}
    \caption{Performance of Qwen3-8B trained with different RL algorithms using our RLTR framework.}
    \label{tab:method}
\end{table}

\noindent \textbf{Different RL Algorithms Results}
As shown in Table~\ref{tab:method}, all three RL algorithms, PPO~\cite{schulman2017proximal}, GRPO~\cite{shao2024deepseekmath}, and REINFORCE++~\cite{hu2025reinforceefficientrlhfalgorithm}, that leverage our RLTR framework, yield consistent performance improvements compared to end-to-end RL using PPO, thereby demonstrating the robustness of our framework. The primary training metrics for each algorithm are reported in detail in Appendix ~\ref{app:rl_metric}.

\subsection{Experiment on Reward Function}
\label{sec:reward_func}
To evaluate the accuracy of our tool-use completeness reward function, we sampled 925 cold-start Planner trajectories and their corresponding final answers, generated by the original Qwen3-8B as the Summarizer, and manually annotated them for correctness. These samples were then evaluated by Qwen3-235B-A22B using both the conventional final answer-based reward and our proposed tool-use completeness reward. The results were compared for alignment with human annotations.

\begin{table}[ht]
\centering
\begin{tabular}{cccc}
\hline
\toprule
\textbf{Reward} & \textbf{ACC}  & \textbf{F1} \\
\hline
Answer & 65.30 & 76.17 \\
Tool-Use Comp. & \textbf{74.59} & \textbf{84.64} \\
\hline
\toprule
\end{tabular}
\caption{Classification results on manually labeled samples using different reward functions. We convert the output values of the reward function $\{0,1\}$ into predicted labels to compute classification metrics.}
\label{tab:reward}
\end{table}

As shown in Table~\ref{tab:reward}, our tool-use completeness reward surpasses the answer reward in both accuracy and F1 score, indicating that it provides more reliable sample evaluation and leads to more stable and effective training.

\section{Related Works}
The development of LLM based agents has witnessed significant advancements, enabling complex task automation by interacting with external tools and environments. In this section, we provide a review of the key related works. We begin by discussing the two primary paradigms for constructing LLM agents: prompt-based and fine-tuning-based approaches. Subsequently, we delve into the burgeoning field of agentic reinforcement learning, which aims to overcome the limitations of earlier methods. We examine current RL strategies and highlight the challenges, such as reward hacking, which our proposed method aims to address.

\subsection{LLM Agents}
LLM-based agents can be broadly categorized into prompt-based and fine-tuning-based approaches~\cite{luo2025large,huang2024understanding}. Prompt-based agents rely on in-context learning to guide the LLM in following tool-use paradigms and interacting with the environment~\cite{shen2023hugginggpt,hongmetagpt,suzgun2024meta}. For example, ReAct improves agent performance by prompting the model to first reason and then perform tool calls~\cite{yao2023react}. However, prompt-based methods are heavily dependent on the capabilities of the underlying base model and are difficult to adapt to specific scenarios. In contrast, fine-tuning-based approaches enhance agent capabilities by updating the LLM’s parameters~\cite{li2023api,qiao2024autoact,ruan2023tptu}. Imitation learning is a common method for rapidly improving agent performance; for instance, ToolAlpaca enhances tool-use ability through supervised fine-tuning (SFT) on high-quality tool-use data~\cite{tang2023toolalpaca}. However, imitation learning often struggles to generalize to out-of-distribution (OOD) queries, which has led to increasing interest in reinforcement learning for agent training.

\subsection{Agentic Reinforcement Learning}
Reinforcement learning (RL)~\cite{kaelbling1996reinforcement} optimizes agents by enabling them to interact with the environment and using the obtained rewards as gradient signals. AgentPRM~\cite{choudhury2025process} introduces a process-level reward model that scores entire trajectories to guide agent optimization. However, such reward models are susceptible to reward hacking, making it challenging to provide reliable and effective supervision. Inspired by DeepSeek-R1~\cite{guo2025deepseek}, recent work has shifted focus to using verifiable rewards for agent training~\cite{song2025r1,chen2025learning,wang2025otc}. For example, Search-R1~\cite{jin2025search} assigns rewards based on the correctness of the final answer to optimize the search trajectory. RAGEN~\cite{ragen} further extends this idea by leveraging multi-turn interactions, where the final output is verified and used as reward feedback. This enables self-evolving training and facilitates more effective reasoning. ToRL~\cite{li2025torl} enables agents to autonomously utilize computational tools through reinforcement learning, allowing the model to explore and discover optimal tool-use strategies.

However, in industrial settings, verifiable questions constitute only a small fraction of real-world queries, which significantly limits the applicability of such methods. Therefore, mainstream LLMs, such as DeepSeek-R1~\cite{guo2025deepseek} and Qwen3~\cite{yang2025qwen3}, still employ general reinforcement learning, where a scoring model is used to assign rewards to the final answers. Nevertheless, reward models are prone to reward hacking~\cite{fu2025reward,liu2024rrm}, resulting in inaccurate rewards. Our approach focuses on assigning rewards based solely on the planning process itself, rather than the final answer, thereby mitigating this issue.

\section{Conclusion}
In this paper, we address the key challenges of unreliable rewards and optimization difficulties in end-to-end agent training. We introduced the Reinforcement Learning with Tool-use Rewards (RLTR) framework, which decouples the problem by focusing on a single-objective optimization of the agent's planning component. By leveraging a novel and more reliable reward signal based on tool-use completeness, our approach circumvents the need for verifiable final answers. Our experiments show that RLTR leads to more stable training and improves action performance by 8\%–12\%. This enhancement directly translates to a 5\%–6\% increase in the accuracy of the agent's final responses. Our work offers a novel perspective for agent optimization in industrial applications.

\section*{Limitations}
Since our deployment scenario is primarily in Chinese, we did not conduct experiments in other major languages such as English or French. Additionally, this work mainly focuses on the optimization of agent planning. Therefore, we employed an untrained LLM as the Summarizer and did not specifically investigate optimization strategies for the Summarizer component. In future work, we plan to construct agent datasets that include more languages to enhance the effectiveness of our approach across different linguistic contexts. We also intend to explore methods for optimizing the Summarizer to further improve the overall performance of the agent system.

\bibliography{anthology,custom}
\bibliographystyle{acl_natbib}

\appendix
\section{Training}
\subsection{Training Template}
\label{app:template}
\begin{figure*}[htbp]
    \centering
    \includegraphics[width=1\linewidth]{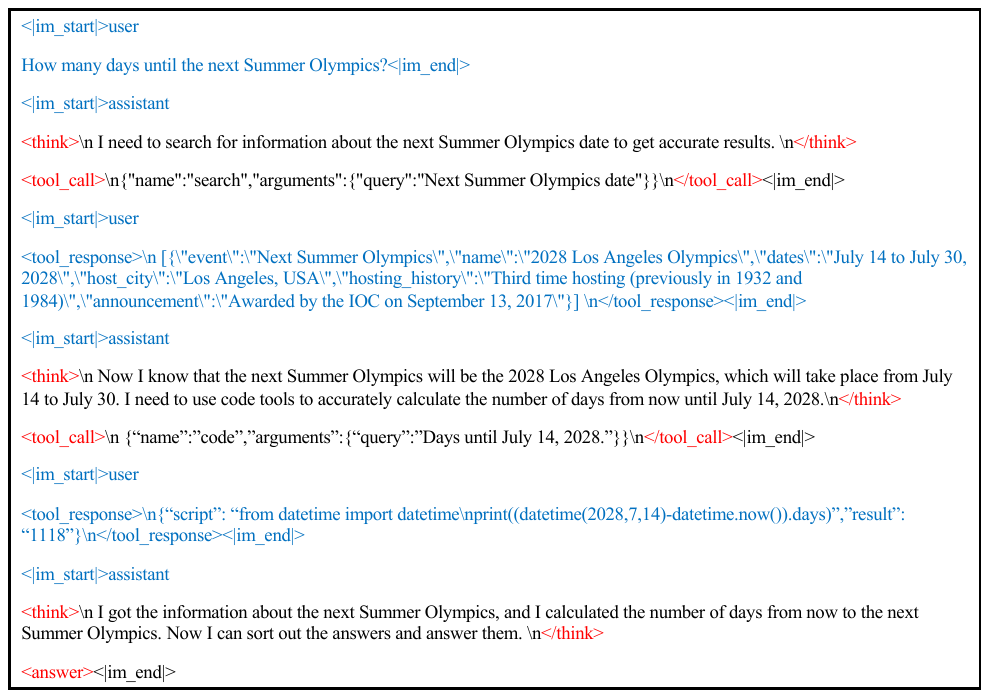}
    \caption{
    Our RLTR training template. The labels in the blue sections are masked during training to prevent loss calculation, ensuring training stability. Unlike full-agent training, the Planner terminates upon outputting to ``<answer>'' and does not generate subsequent responses.
}
    \label{fig:template}
\end{figure*}

As shown in Figure~\ref{fig:template}, we utilize the native inference and tool-use templates from Qwen. For the Planner, we focus training exclusively on the action component and the reasoning phase just before the final answer. We mask the loss from non-answer parts, allowing the Planner to concentrate on optimizing tool calls, thereby stabilizing the training process.
\label{app:train_template}

\begin{figure*}[htbp]
    \centering
    \includegraphics[width=1\linewidth]{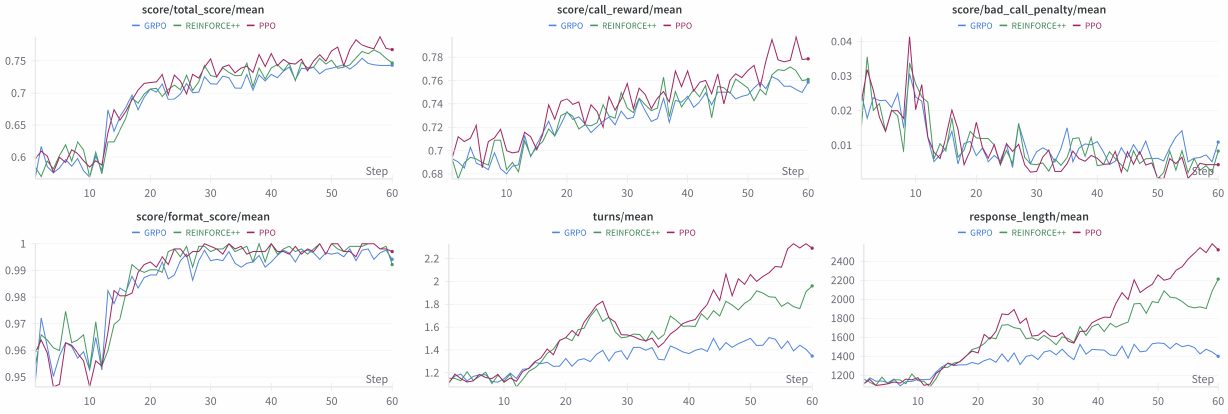}
    \caption{Trends of main metrics for PPO, REINFORCE++, and GRPO algorithms during general reinforcement learning on Qwen3-8B.}
    \label{fig:rl_metric}
\end{figure*}
\subsection{Training Algorithm}
\label{app:rl_alg}
As shown in Algorithm~\ref{alg:rl}, our RL framework supports three common RL algorithms: PPO~\cite{schulman2017proximal}, GRPO~\cite{shao2024deepseekmath}, and REINFORCE++~\cite{hu2025reinforceefficientrlhfalgorithm}. All three algorithms utilize the tool-use completeness we proposed as the primary action reward metric. The process follows a generate-evaluate-optimize cycle until the Planner's policy converges.

\begin{algorithm}[htbp]
\caption{Multi-Turn Reinforcement Learning for Planner Optimization}
\label{alg:rl}

\KwIn{Initial Planner policy $\pi_p$, Reference policy $\pi_{\mathrm{ref}}$, Data distribution $\mathcal{D}$, Set of tools $\mathcal{T}$, KL-divergence weight $\beta$.}
\KwOut{Optimized Planner policy $\pi_p^*$.}

\BlankLine %

Initialize Planner parameters $\theta_p$\;

\While{not converged}{
    
    \tcc{Phase 1: Generate Trajectories}
    Sample a batch of queries $\{x_i\}_{i=1}^B$ from $\mathcal{D}$\;
    Generate a batch of tool-use trajectories $\{\tau_i\}_{i=1}^B$ where $\tau_i \sim \pi_p(\cdot | x_i; \mathcal{T})$\;
    
    \BlankLine
    \tcc{Phase 2: Evaluate and Compute Rewards}
    \For{each trajectory $\tau_i$ in the batch}{
        \If{trajectory format of $\tau_i$ is invalid}{
            $R_{\mathrm{total}, i} \leftarrow -1$\;
        }
        \Else{
            
            \tcp{Calculate completeness reward using a verification LLM}
            $R_{\mathrm{comp}, i} \leftarrow \frac{1}{N}\sum_{j=1}^{N} \gamma_j(\tau_i)$\;
            
            \tcp{Calculate rule-based penalties}
            $R_{\mathrm{repeat}, i} \leftarrow -\lambda \sum_{t=2}^{T} \mathbb{I}(a_t = a_{t-1})$\;
            $R_{\mathrm{error}, i} \leftarrow -\mu \sum_{t=1}^{T} \mathbb{I}(a_t \notin \mathcal{A}^*_t)$\;
            $R_{\mathrm{rule}, i} \leftarrow R_{\mathrm{repeat}, i} + R_{\mathrm{error}, i}$\;
            
            $R_{\mathrm{total}, i} \leftarrow R_{\mathrm{comp}, i} + R_{\mathrm{rule}, i}$\;
        }
    }
    
    \BlankLine
    \tcc{Phase 3: Optimize Policy}
    Update the Planner's parameters $\theta_p$ by optimizing the objective:\;
    $\arg\max_{\pi_{p}} \mathbb{E}_{x\sim\mathcal{D}, a\sim\pi_{p}}\left[R_{\mathrm{total}}(x,a)\right] - \beta D_{\mathrm{KL}}\left(\pi_{p}(a|x;\mathcal{T}) \,\|\, \pi_{\mathrm{ref}}(a|x;\mathcal{T})\right)$\;
}

\BlankLine
\KwRet{$\pi_p^* \leftarrow \pi_p$}\;
\end{algorithm}

\subsection{Implementation Details}
\label{app:detail}
In the cold-start phase, we utilize Qwen3-32B~\cite{yang2025qwen3} as the teacher LLM for knowledge distillation. We adopt verl~\cite{sheng2024hybridflow} as our reinforcement learning framework and apply its recommended hyperparameters for PPO, GRPO, and REINFORCE++. All RL training experiments are conducted on $2\times8$ H20 GPUs. 
\subsection{RL Training Metric}
\label{app:rl_metric}
As shown in Figure~\ref{fig:rl_metric}, both the tool calling reward and the total reward exhibit an upward trend as the number of RL training steps increases, while the calling error penalty gradually decreases. Simultaneously, the number of calls and the response length also increase, indicating that the Planner progressively improves its tool calling completeness and thereby acquires more comprehensive information. Additionally, we observed that, compared to PPO and REINFORCE++, GRPO did not demonstrate significant growth in either the number of calls or the response length. This phenomenon can be attributed to the unique Group Normalization mechanism employed by GRPO. Specifically, the normalization mechanism in GRPO shifts the optimization objective toward achieving the relatively highest score among multiple responses generated from the same prompt. As a result, the direct correlation between response length and the final reward signal is reduced, meaning the model no longer needs to rely on increasing the number of calls for optimization. Instead, it focuses on refining the tool calling parameters to initiate more accurate calls, leading to a more gradual increase in both response length and the number of calls.

\section{Prompt}
\label{app:prompt_check}
\label{app:answer_prompt}
\label{app:prompt_check}

\begin{figure}[t]
    \centering
    \includegraphics[width=1.0\linewidth]{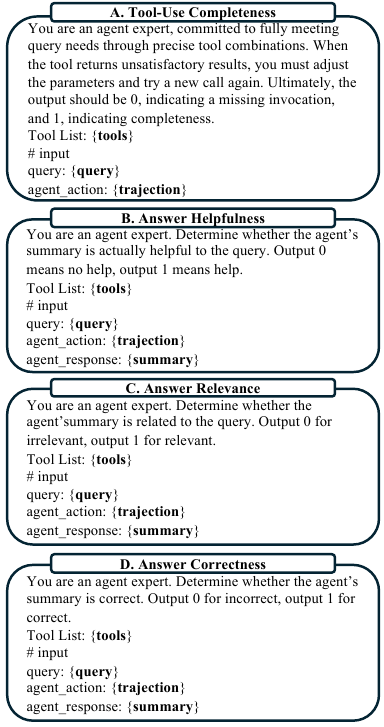}
    \caption{(A) tool-use completeness, (B) answer helpfulness, (C) answer relevance, and (D) answer correctness evaluation prompts.}
    \label{fig:prompt}
\end{figure}

\section{Agent Environment}
\label{app:env}
\subsection{Tools}
\label{app:tool}
\begin{itemize}
    \item \textbf{Search} A real-time search tool using the Sogou API~\footnote{https://data.open.sogou.com/}, tailored for Chinese language scenarios.
    \item \textbf{Code} A tool that converts natural language instructions into Python code and executes it within a secure sandbox.
\end{itemize}

\begin{figure*}[t]
    \centering
    \includegraphics[width=0.8\linewidth]{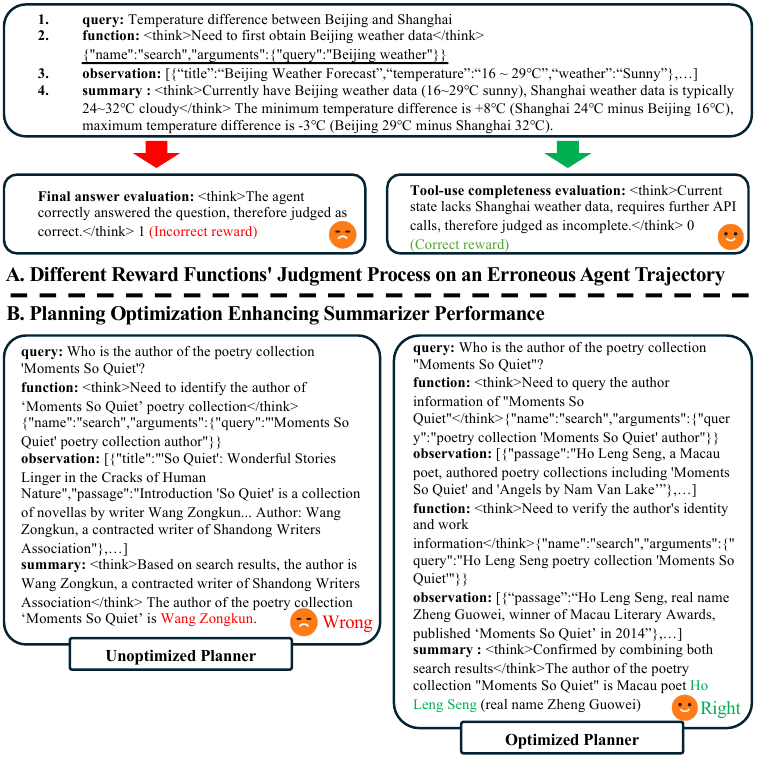}
    \caption{Examples illustrating the evaluation of different reward criteria and the impact of Planner optimization on summary generation.}
    \label{fig:case}
\end{figure*}

\section{Case Study}

\noindent \textbf{Tool-Use Completeness Reward is More Accurate than Final Answer Correctness Reward.} As shown in Figure~\ref{fig:case} (A), the task is to calculate the temperature difference between Beijing and Shanghai. To solve this, the agent must use a search tool to retrieve weather data for both cities and then employ a code tool to compute the temperature difference. When the reward is based solely on the correctness of the final answer, the agent is incorrectly given a positive reward, as it appears to have answered the question correctly. However, the weather data for Shanghai is fabricated by the agent. In contrast, the tool-use completeness reward successfully identifies the missing steps—specifically, searching for Shanghai's weather data and calculating the final temperature difference—and thus assigns the correct negative reward for the incomplete trajectory.

\noindent \textbf{Optimized Planner Can Improve Summarizer Performance.} As shown in Figure~\ref{fig:case} (B), for a complex factual query, the unoptimized Planner fails to make complete tool calls, resulting in insufficient information for the Summarizer to generate an accurate summary, and ultimately leads to an incorrect response. In contrast, the optimized Planner initiates further searches when information is incomplete, eventually obtaining accurate data. This enables the Summarizer to organize the answer based on reliable information, thereby producing a correct response.

\end{document}